\documentclass[10pt,twocolumn,letterpaper]{article}
\usepackage[pagenumbers]{cvpr}
\usepackage{graphicx}
\usepackage{amsmath,amssymb}
\usepackage{booktabs,tabularx,array,multirow}
\usepackage{xcolor,xspace}
\usepackage{enumitem}
\usepackage{microtype}
\definecolor{realizeblue}{RGB}{32,88,144}
\definecolor{realizeteal}{RGB}{0,119,126}
\definecolor{pending}{RGB}{148,66,36}
\newcommand{\method}{MakoBORN\xspace}
\newcommand{\term}[1]{\textbf{\textit{#1}}}   

\newcolumntype{Y}{>{\raggedright\arraybackslash}X}

\setlist[itemize]{leftmargin=*,itemsep=2pt,topsep=3pt}
\setlist[enumerate]{leftmargin=*,itemsep=2pt,topsep=3pt}
\usepackage{pifont}
\usepackage{fontawesome5}
\usepackage{sansmath}
\definecolor{yesgreen}{RGB}{0,140,60}
\definecolor{nored}{RGB}{200,30,30}
\usepackage{colortbl}
\usepackage[most]{tcolorbox}
\usetikzlibrary{arrows.meta,calc,shadows,shapes.arrows,shapes.symbols}
\newtcolorbox{questionbox}[1][]{enhanced, breakable=false, colback=yesgreen!6, colframe=yesgreen,
  boxrule=0pt, leftrule=2pt, arc=0pt, outer arc=0pt, left=5pt, right=5pt, top=3pt, bottom=3pt,
  before skip=6pt, after skip=4pt, fontupper=\small, #1}
\newtcolorbox{runinframe}[1][]{enhanced, breakable=false, colback=black!3, colframe=black!25,
  boxrule=0.5pt, arc=0pt, outer arc=0pt, left=6pt, right=6pt, top=4pt, bottom=4pt,
  before skip=6pt, after skip=6pt, #1}
\newcommand{\runinbox}[2]{\begin{runinframe}\textbf{#1:}\ #2\end{runinframe}}
\newcommand{\qbox}[2]{\begin{questionbox}\textbf{#1}\par\vspace{1pt}#2\end{questionbox}}
\newcommand{\qpart}[4][]{\subsubsection{#2}\if\relax\detokenize{#1}\relax\else\label{#1}\fi\nopagebreak[4]%
  \begin{questionbox}[before skip=1pt]\textbf{#3}\par\vspace{1pt}#4\end{questionbox}\noindent\ignorespaces}
\makeatletter
\def\cvprsubsubsection{\@startsection {subsubsection}{3}{\z@}
   {-6pt plus -2pt minus -2pt}{3pt} {\tenbf\raggedright}}
\makeatother
\usepackage[normalem]{ulem}
\newcommand{\lvl}[1]{\uline{\textit{#1}}.\ }
\providecommand{\doi}[1]{\href{https://doi.org/#1}{doi:\nolinkurl{#1}}}

\definecolor{cvprblue}{rgb}{0.21,0.49,0.74}
\usepackage[pagebackref,breaklinks,colorlinks,allcolors=cvprblue]{hyperref}
\def\confName{CVPR}
\def\confYear{2027}
\title{Born for the Task:\\ Rethinking Quantum Chemistry with Foresighted Agentic Computing}
\hypersetup{pdftitle={Born for the Task: Rethinking Quantum Chemistry with Foresighted Agentic Computing},pdfauthor={Haozhi Han, Junjie Yang, Ruge Zhang, Zihe Huang, Pengyi Xu, Bo Zhang, Xianyi Zhang, Yunquan Zhang, Ting Cao, Yunxin Liu, Yifeng Chen, Kun Li}}
\newcommand{\paperAuthorNames}{%
\mbox{Haozhi Han\textsuperscript{1,5,*}},
\mbox{Junjie Yang\textsuperscript{2,*}},
\mbox{Ruge Zhang\textsuperscript{3,5}},
\mbox{Zihe Huang\textsuperscript{1,5}},
\mbox{Pengyi Xu\textsuperscript{4}},
\mbox{Bo Zhang\textsuperscript{4}},
\mbox{Xianyi Zhang\textsuperscript{6}},
\mbox{Yunquan Zhang\textsuperscript{3}},
\mbox{Ting Cao\textsuperscript{5}},
\mbox{Yunxin Liu\textsuperscript{5}},
\mbox{Yifeng Chen\textsuperscript{1}},
and \mbox{Kun Li\textsuperscript{5,\dag}}}
\newcommand{\paperAffiliations}{%
\textsuperscript{1}School of Computer Science, Peking University\\
\textsuperscript{2}Moore Threads\\
\textsuperscript{3}University of Chinese Academy of Sciences\\
\textsuperscript{4}Michigan Technological University\\
\textsuperscript{5}Institute for AI Industry Research (AIR), Tsinghua University\\
\textsuperscript{6}PerfXLab Technology
}
\newcommand{\paperDate}{November 18, 2026}
\newcommand{\paperEqualContribution}{\textsuperscript{*}Equal contribution.}
\newcommand{\paperCorrespondingAuthor}{\textsuperscript{\dag}Corresponding author.}
\newcommand{\paperEmail}{\href{mailto:likun@air.tsinghua.edu.cn}{\textcolor{FrontTitleBlue}{likun@air.tsinghua.edu.cn}}}
\author{\paperAuthorNames}

\definecolor{FrontBoxBlue}{HTML}{F0F4F8}
\definecolor{FrontTitleBlue}{HTML}{2F5DA8}
\definecolor{BornGradientStart}{HTML}{701426}
\definecolor{BornGradientMiddle}{HTML}{8F1D2C}
\definecolor{BornGradientEnd}{HTML}{B2474D}
\newsavebox{\bornTitleBox}
\newsavebox{\bornGradientBox}
\newcommand{\bornGradient}{%
  \begingroup
  \sbox{\bornTitleBox}{Born}%
  \sbox{\bornGradientBox}{%
  \leavevmode\pdfliteral {q}%
  \rlap{\pdfliteral {7 Tr}\usebox{\bornTitleBox}\pdfliteral {0 Tr}}%
  \raisebox{-\dp\bornTitleBox}{%
    \tikz\shade[left color=BornGradientStart,right color=BornGradientEnd,middle color=BornGradientMiddle]
      (0,0) rectangle (\wd\bornTitleBox,\dimexpr\ht\bornTitleBox+\dp\bornTitleBox\relax);}%
  \pdfliteral {Q}}%
  \pdfxform\bornGradientBox
  \leavevmode\pdfrefxform\pdflastxform
  \endgroup
}
\newtcolorbox{frontmatterbox}{
  enhanced, width=\textwidth,
  colback=FrontBoxBlue, colframe=FrontBoxBlue,
  boxrule=0pt, arc=12pt, outer arc=12pt,
  left=8mm, right=8mm, top=7mm, bottom=7mm,
  before skip=0pt, after skip=5mm
}
\makeatletter
\newcommand{\makepaperfrontmatter}[2]{%
  \twocolumn[{%
    \begin{@twocolumnfalse}
    \vspace*{-9mm}%
    \begin{frontmatterbox}
      {\raggedright\fontfamily{phv}\fontsize{18.5}{22}\selectfont\bfseries
        {\itshape\color{FrontTitleBlue}\bornGradient{} for the Task:} Rethinking Quantum Chemistry\\
        with Foresighted Agentic Computing\par}
      #1%
      \vspace{0.38cm}
      {\raggedright\fontfamily{phv}\fontsize{10.4}{13.4}\selectfont\bfseries
        \paperAuthorNames\par}
      \vspace{0.22cm}
      {\raggedright\fontsize{9.7}{11.9}\selectfont\normalfont
        \paperAffiliations\par
        \vspace{0.10cm}\paperEqualContribution\quad\paperCorrespondingAuthor\par}
      #2%
      \par\vspace{0.38cm}
      \noindent\begin{minipage}[b]{0.56\linewidth}
      \raggedright\fontsize{9.7}{12.8}\selectfont\normalfont
        {\fontfamily{phv}\selectfont\bfseries Date:} \paperDate\par
        {\fontfamily{phv}\selectfont\bfseries Email:} \paperEmail\par
      \end{minipage}\hfill
      \begin{minipage}[b]{0.42\linewidth}
        \raggedleft
        \raisebox{-0.0183in}[0.30in][0pt]{\includegraphics[height=0.382in]{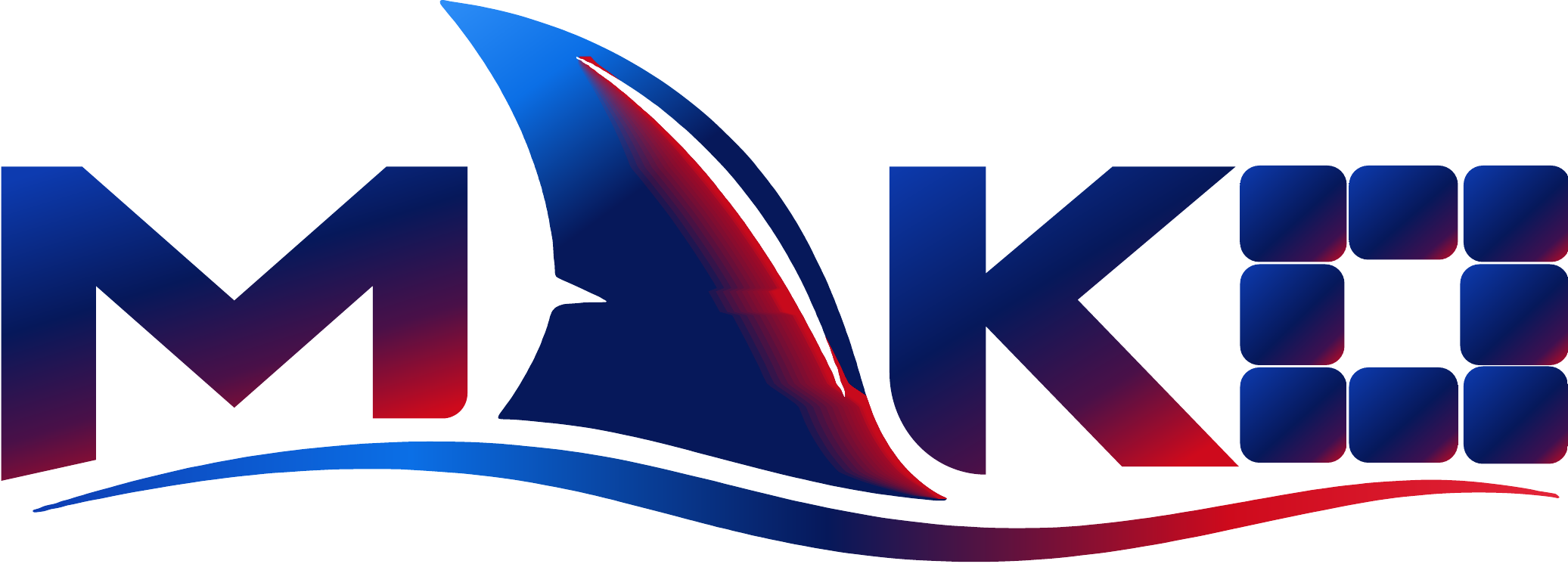}}\hspace{0.16in}%
        \href{https://air.tsinghua.edu.cn/}{\raisebox{0.01in}{\includegraphics[height=0.28in]{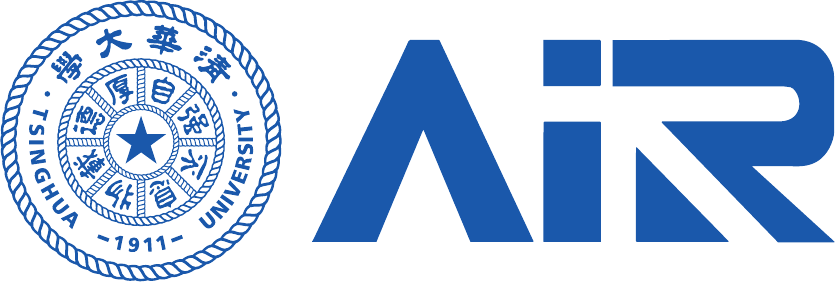}}}\par
      \end{minipage}
    \end{frontmatterbox}
    \end{@twocolumnfalse}
  }]%
}
\makeatother

\makeatletter
\def\cvprsection{\@startsection{section}{1}{\z@}
  {-8pt plus -2pt minus -1pt}{5pt}{\large\bf}}
\def\cvprsubsection{\@startsection{subsection}{2}{\z@}
  {-6pt plus -1pt minus -1pt}{4pt}{\elvbf}}
\def\cvprsubsubsection{\@startsection{subsubsection}{3}{\z@}
  {-5pt plus -1pt minus -1pt}{2pt}{\tenbf\raggedright}}
\makeatother

\begin{document}
\makepaperfrontmatter{}{%
  \vspace{0.44cm}
  {\fontsize{9.8}{12.1}\selectfont\normalfont
    \setlength{\parindent}{0pt}%
    \renewenvironment{abstract}{\ignorespaces}{\par}%
    \begin{abstract}
Quantum chemistry relies on general-purpose software, whose generality leaves many large-scale calculations beyond practical resource budgets. Coding agents make task-specific programs cheap to write, but not efficient ones easy to find: whether a computational strategy is accurate and fast enough becomes clear only after it actually runs at the target scale. We introduce \method, a foresighted agentic computing framework that synthesizes high-performance quantum chemistry programs on demand. \method anticipates the behavior of each strategy before execution, extrapolating its error from representative small systems and predicting its runtime from the target-scale workload and measured hardware throughputs. Building on these predictions, \method unfolds the preset schemes of existing packages into independently configurable algorithmic components, selects the fastest strategy within the error tolerance from the predicted accuracy--runtime Pareto frontier, and guides coding agents to compile it with component-level performance predictions. For a Hartree--Fock calculation with 50{,}024 basis functions, \method finishes in 1.41\,h on a single CPU node, 77$\times$ faster than the estimated runtime of a general-purpose baseline and still 9.8$\times$ faster with development time included. On a single GPU, \method completes the ubiquitin calculation in 2.1\,h, the same order of runtime as the GPU-accelerated package Mako on 64 GPUs, and shortens the B-DNA duplex calculation from an estimated two months to 8.1\,h (173$\times$).
\end{abstract}

  }%
}
\begingroup
\renewcommand{\thefootnote}{}%
\footnotetext{This work is supported by PerfXLab Technology and Wuxi Research Institute of Applied Technologies, Tsinghua University under Grant 20242001120.}%
\endgroup
\section{Introduction}
\suppressfloats[t]
\begin{figure}[t]
\centering
\input{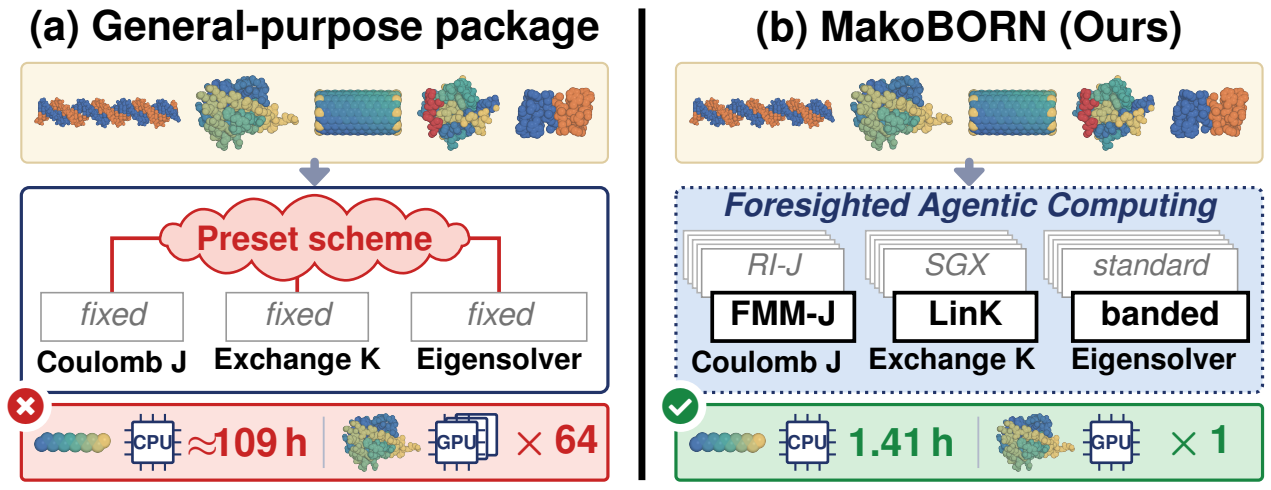}
\caption{\textbf{General-purpose package vs.\ \method: one preset scheme for every task vs.\ a program born for each task.}}
\label{fig:paradigm}
\end{figure}

Quantum chemistry resolves the structure and properties of matter at the electronic level by solving the electronic Schrödinger equation~\cite{helgaker2000book}. Its numerical solution relies on mature software packages developed and refined by domain experts over decades~\cite{neese2025orca6,epifanovsky2021qchem,sun2020pyscf,wu2025gpu4pyscf}. Given the high cost of software development and numerical validation, these packages combine broadly applicable methods with carefully optimized general-purpose implementations, delivering reliable results across diverse molecules, scientific objectives, and computing environments. 

However, this generality can limit computational efficiency for demanding quantum chemistry tasks. Each calculation fixes its molecule, basis set, accuracy requirement, and hardware, yet general-purpose implementations offer only a few preset algorithms and settings that cannot exploit them. This limitation is particularly pronounced for large molecular systems of scientific interest, where the time or memory required by general-purpose implementations can exceed practical resource budgets~\cite{goedecker1999linear,kussmann2013linear}. Consequently, the quantum chemistry calculations needed to investigate these systems can remain beyond practical reach.

Recent advances in coding agents have opened new opportunities for quantum chemistry through automated code generation and optimization~\cite{chen2021codex,yang2024sweagent,novikov2025alphaevolve}. By lowering development costs, these agents make it increasingly feasible to synthesize programs tailored to individual quantum chemistry tasks~\cite{hagai2026ladeq,shen2026gamess,chow2026optimize}, rather than relying solely on preset general-purpose implementations. Such programs can exploit what each task fixes, bringing previously infeasible calculations within practical reach.

Yet while coding agents make programs easy to write, they do not make efficient programs easy to find, for three reasons.
\textit{Unproductive search:} beyond preset schemes, algorithms and settings combine into a vast, unmapped space, so agents waste the budget on unpromising strategies.
\textit{Expensive feedback:} a strategy's accuracy and runtime are revealed only by running it, yet target-scale runs are costly and small-scale runs mislead, so agents cannot judge strategies cheaply.
\textit{Blind optimization:} small-scale timings also hide target-scale bottlenecks, so agents cannot tell where optimization pays off.
This raises a central question: \textit{How can coding agents build the program best suited to a task before running it at the target scale?}

We answer this question with \method\footnote{\method continues our Mako series~\cite{han2025mako,han2026makoxc}. Mako, named after the world's fastest shark, reflects our pursuit of pushing quantum chemistry computation to its performance limits. BORN captures our vision of programs \emph{born for the task} through foresighted synthesis and honors Max Born, whose Born--Oppenheimer approximation underpins modern electronic-structure calculations.}, a foresighted agentic computing framework that guides the on-demand synthesis of high-performance programs through predictive reasoning, extending the computational frontier of quantum chemistry beyond existing general-purpose software. Our key insight is that the feasibility and performance potential of a computational strategy can be anticipated before it is executed. Specifically, its numerical error can be estimated by analyzing representative small systems and extrapolating to the target~\cite{kohn1996density,prodan2005nearsightedness}, while its runtime can be predicted by combining the target-scale computational workload with throughputs measured by hardware microbenchmarks~\cite{williams2009roofline,volkov2008benchmarking}. This allows agents to guide strategy selection and program synthesis through foresighted decisions, without relying on costly target-scale trial and error.

\method puts this insight into practice through three interlocking stages.
\textbf{Component-Wise Unfolding} decomposes preset schemes in general-purpose software into independently configurable algorithmic components and recombines them across components and computational stages, exposing a structured design space with potentially efficient strategies unavailable in existing packages.
\textbf{Target-Scale Foresight} reuses component-level analyses to accurately predict numerical errors and runtimes at the target scale at low cost, constructing a predicted accuracy--runtime Pareto frontier and selecting the fastest strategy expected to satisfy the accuracy constraint.
\textbf{Agent-Driven Compilation} exploits the task-specific computational structure for ahead-of-time specialization and uses component-level performance predictions to diagnose bottlenecks, guiding agents toward targeted, hardware-aware optimization and iterative validation.

We evaluate \method on large-scale quantum chemistry tasks, demonstrating its ability to extend the computational frontier beyond general-purpose software. For a Hartree--Fock calculation with 50{,}024 basis functions, \method completes the task in just 1.41\,h on a single CPU node, while some general-purpose packages fail to complete it. Even a general-purpose baseline capable of running the task is estimated to require 109\,h, corresponding to a 77$\times$ computational speedup and a 9.8$\times$ end-to-end speedup including development time. For ubiquitin, \method reduces the GPU requirement from 64 to just one while maintaining comparable runtime.

\textbf{More broadly, we envision a paradigm shift in scientific computing: from adapting scientific problems to existing programs to building programs born for each task—expanding not just computational efficiency, but the boundaries of what we can compute.}
\begin{figure*}[t]
\centering
\input{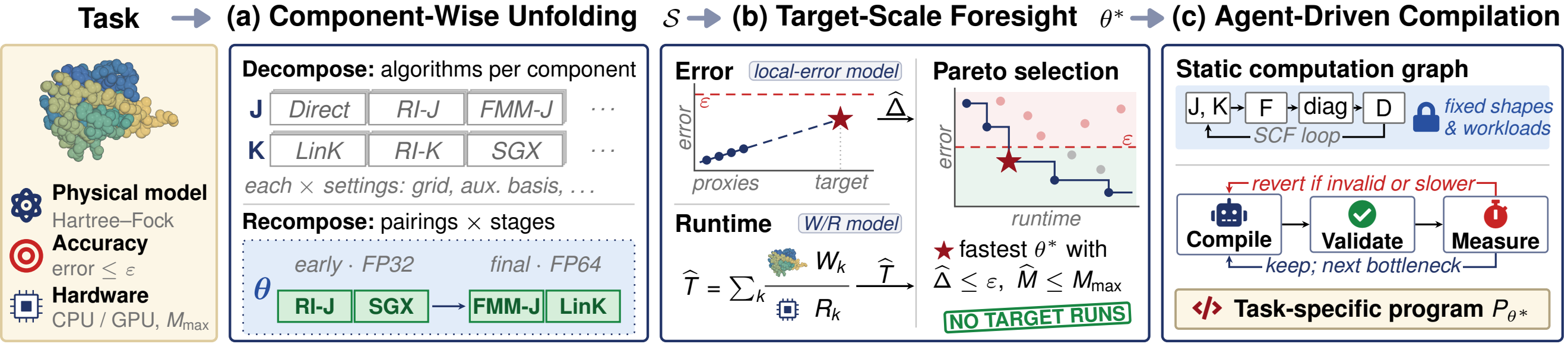}
\caption{\textbf{Overview of \method.} Given a task, \method (a)~unfolds package schemes into a strategy space $\mathcal{S}$, where one strategy $\theta\in\mathcal{S}$ is one pairing per SCF stage; (b)~predicts the target-scale error and runtime of every strategy from proxies and microbenchmarks and selects the fastest $\theta^\ast$ within $\varepsilon$; and (c)~has coding agents compile $\theta^\ast$ into a task-specific program.}
\label{fig:overview}
\end{figure*}
\section{Related Work}

\textbf{Quantum Chemistry Software and Algorithms.}
ORCA, Q-Chem, Psi4, NWChem, GAMESS, PySCF, and GPU4PySCF~\cite{neese2022orca,epifanovsky2021qchem,smith2020psi4,apra2020nwchem,barca2020gamess,sun2020pyscf,wu2025gpu4pyscf} accelerate SCF with fast Coulomb (RI-J, FMM)~\cite{vahtras1993integral,eichkorn1995auxiliary,white1994cfmm} and exchange builds (LinK, RI-K variants, SGX/COSX)~\cite{ochsenfeld1998link,weigend2002rihf,polly2004local,manzer2015parik,friesner1985pseudospectral,neese2009cosx}, but expose them as a few preset schemes with parameters reliable across molecules. TeraChem, EXESS, Mako, and MakoXC port SCF to GPUs and AI accelerators~\cite{seritan2021terachem,barca2020scaling,barca2021faster,galvez2023toward,palethorpe2024advanced,han2025mako,han2026makoxc}\\
\textbf{Agentic Scientific Programming.}
AlphaEvolve evolves code under automated evaluators~\cite{novikov2025alphaevolve}, and KernelBench measures the correctness and efficiency of agent-generated GPU kernels~\cite{ouyang2025kernelbench}; cuPilot and KernelPro add roofline reasoning or hardware profiling~\cite{chen2025cupilot,gai2026kernelpro}. Scientific agents improve scientific machine-learning workflows~\cite{zhong2026agent4sci}, LADeQ generates quantum-chemical approximations with error control~\cite{hagai2026ladeq}, and agents have modernized the production code GAMESS~\cite{shen2026gamess}. Agent harnesses run multi-step simulations autonomously on existing packages~\cite{meng2026fermilink,ding2026openclaw}, and the worker--controller loop of FermiLink optimizes the J/K and TDDFT kernels of PySCF, accepting changes on measured runtime and accuracy~\cite{chow2026optimize}. These agents improve code or approximations from measured feedback. \method instead predicts target-scale error and runtime before implementation, so that development effort goes only to the strategy and components that can pay off.
\section{Problem Formulation}
\label{sec:problem}

\textbf{Tasks, Strategies, and Programs.}
We take Gaussian-basis Hartree--Fock (HF), solved by self-consistent-field (SCF) iteration~\cite{roothaan1951new,hall1951molecular}, as the concrete setting. A task \(x\) specifies the molecule, electronic state, HF formulation, orbital basis, error tolerance \(\varepsilon\), and target hardware. We separate what is computed from how it is computed. A \emph{computational strategy} \(\theta\in\mathcal{S}\) fixes the algorithm and numerical parameters of every component of the calculation and how they change over the SCF run (Sec.~\ref{sec:compose}); it determines the numerical error \(\Delta(\theta;x)\) relative to a reference solution without further numerical approximation. A \emph{program} \(P_\theta\) implements \(\theta\) on the target hardware; it determines the runtime \(T(P_\theta;x)\) and the peak memory \(M_{\mathrm{peak}}(P_\theta;x)\). A general-purpose package fixes one generic program for all tasks and restricts strategies to a few preset schemes \(\mathcal{S}_{\mathrm{pkg}}\subset\mathcal{S}\). We instead treat both the strategy and the program as decision variables and build a program for each task.

\textbf{Objective.}
Given \(x\), the memory limit \(M_{\max}\) of its hardware, and a development budget \(B\), we minimize the total time to solution:
\begin{equation}
\label{eq:problem}
\begin{aligned}
\min_{\theta\in\mathcal{S},\,P_\theta} \quad &
T_{\mathrm{dev}}+T(P_\theta;x)\\
\text{s.t.}\quad &
\Delta(\theta;x)\leq\varepsilon,\quad
M_{\mathrm{peak}}(P_\theta;x)\leq M_{\max},\\
& T_{\mathrm{dev}}\leq B,
\end{aligned}
\end{equation}
where \(T_{\mathrm{dev}}\) covers all analysis, code generation, trial runs, and validation. We state the single-run case; additional production runs only amortize \(T_{\mathrm{dev}}\). When no scheme in \(\mathcal{S}_{\mathrm{pkg}}\), run by the package's generic program, meets \(\varepsilon\) within \(M_{\max}\) and practical time, the task lies beyond the practical reach of general-purpose software; enlarging the space to \(\mathcal{S}\) and specializing \(P_\theta\) to the task can bring it within reach.

\textbf{Decomposition.}
Let \(\widehat{T}(\theta;x)\) be the predicted runtime of the uncompiled program, which implements \(\theta\) with general-purpose kernels and solvers (Sec.~\ref{sec:search}). Then
\begin{equation}
\label{eq:decomp}
T_{\mathrm{dev}}+T
=
\underbrace{\widehat{T}}_{\substack{\text{strategy}\\\text{cost}}}
-
\underbrace{\bigl(\widehat{T}-T\bigr)}_{\substack{\text{compilation}\\\text{gain}}}
+
\underbrace{T_{\mathrm{dev}}}_{\substack{\text{development}\\\text{cost}}}.
\end{equation}
The three terms are set by the choice of \(\theta\), the specialization of \(P_\theta\), and the effort spent within \(B\); Secs.~\ref{sec:predict}--\ref{sec:e2e} measure them in turn.

\textbf{Challenges.}
The objective and constraints of Eq.~(\ref{eq:problem}) are defined at the target scale, yet evaluating them directly requires target-scale runs that take hours to days. Each of the three decisions, where to search, which strategy to select, and how to compile it, therefore faces one challenge.
(1)~\textit{Unproductive search:} beyond \(\mathcal{S}_{\mathrm{pkg}}\), choices combine across components and SCF stages into a vast, unmapped space, which must be organized for component-wise analysis (Sec.~\ref{sec:compose}).
(2)~\textit{Expensive feedback:} \(\Delta\) and \(T\) are revealed only by running, and small systems mislead because errors and runtime scaling change with size; both must be predicted at the target scale without running there (Sec.~\ref{sec:search}).
(3)~\textit{Blind optimization:} the runtime shares of components change with size, so small-scale timings hide target-scale bottlenecks; compilation must target the dominant components while keeping \(\Delta\leq\varepsilon\) (Sec.~\ref{sec:compile}).

\section{The \method Framework}
\label{sec:method}

\method addresses the three challenges in three stages (Fig.~\ref{fig:overview}). Component-Wise Unfolding builds the strategy space \(\mathcal{S}\) (Sec.~\ref{sec:compose}); Target-Scale Foresight predicts the error, runtime, and peak memory of every strategy at the target scale and selects \(\theta^\ast\), together with the predicted runtime \(\widehat{T}_i\) of each component (Sec.~\ref{sec:search}); and Agent-Driven Compilation turns \(\theta^\ast\) into a task-specific program \(P_{\theta^\ast}\) (Sec.~\ref{sec:compile}). In terms of Eq.~(\ref{eq:decomp}), the first two stages minimize the strategy cost, the third earns the compilation gain, and predictions in place of target-scale trial runs keep the development cost within \(B\).

\subsection{Component-Wise Unfolding}
\label{sec:compose}

Existing packages offer strategies only as preset schemes \(\mathcal{S}_{\mathrm{pkg}}\), such as RIJCOSX in ORCA. Each scheme bundles three levels of choice: the algorithm for each component of the calculation, how these algorithms are paired, and how the pairing is scheduled over the SCF run. Because the fastest choice at each level depends on the molecule, accuracy, and hardware, fixing them jointly for every task forfeits the strategies a task needs. \method unbundles these levels, \emph{algorithms}, \emph{pairings}, and \emph{schedules}, and recomposes them subject to compatibility; the resulting space contains every package scheme as a special case.

\textbf{Knowledge-Based Decomposition: Algorithms.}
Each SCF iteration of restricted HF (RHF) consists of components \(i\in\mathcal{I}\): Coulomb (J) construction, exchange (K) construction, Fock-matrix diagonalization, and the density update. Domain knowledge identifies these components as separately replaceable and supplies the candidate algorithms \(\mathcal{A}_i\) of each; this decomposition is specific to the method but not to the task, so it is set up once for HF and reused across tasks.
J may be built directly, by the resolution of the identity (RI-J), or with the fast multipole method (FMM-J); K directly, with linear-scaling density-based screening (LinK), by RI with global, local, or pair-atomic fitting (RI-K, local RI-K, PARI-K), or by seminumerical quadrature (SGX, including chain-of-spheres exchange, COSX). J and K carry the main approximations; diagonalization and the density update~\cite{pulay1980diis,pulay1982diis} enter the strategy through their precision and convergence thresholds.
Each algorithm \(a\in\mathcal{A}_i\) takes numerical parameters, such as the auxiliary basis, quadrature grid, screening threshold, and floating-point precision~\cite{weigend2006jkfit,lebedev1976quadratures,haser1989improvements,luehr2011dynamic}, with \(Q_{i,a}\) settings; since they set the accuracy--cost trade-off, they are part of the choice, not fixed defaults. An \emph{algorithm choice} for component \(i\) is one algorithm with one setting, drawn from \(\mathcal{C}_i\) with \(|\mathcal{C}_i|=\sum_{a\in\mathcal{A}_i}Q_{i,a}\).
A choice belongs to the strategy if it changes the approximation, and to the program otherwise: integral handling (direct or stored), incremental Fock builds, and integral kernels~\cite{almlof1982direct,obara1986efficient} change only the cost and are deferred to compilation (Sec.~\ref{sec:compile}).

\textbf{Combinatorial Composition: Pairings and Schedules.}
A \emph{pairing} selects one algorithm choice for every component of an SCF stage; any J algorithm can be paired with any K algorithm.
A \emph{schedule} follows an SCF protocol \(\pi\in\Pi\), which divides the iterations into stages \(\mathcal{T}(\pi)\), optionally ending with a post-SCF energy re-evaluation, and assigns each stage its own pairing, e.g., looser thresholds and lower precision early and tighter ones near convergence~\cite{luehr2011dynamic}.
A computational strategy is a complete schedule, i.e., one pairing per stage. Before compatibility constraints, the strategy space has size
\begin{equation}
|\mathcal{S}_{\mathrm{raw}}|
=
\sum_{\pi\in\Pi}
\prod_{s\in\mathcal{T}(\pi)}
\prod_{i\in\mathcal{I}_s}
|\mathcal{C}_i|,
\end{equation}
where \(\mathcal{I}_s\subseteq\mathcal{I}\) are the components active in stage \(s\); the nested terms mirror algorithms (\(|\mathcal{C}_i|\)), pairings (\(\prod_{i}\)), and schedules (\(\prod_{s}\), \(\sum_{\pi}\)).
Algorithmic dependencies and cross-stage compatibility, such as tightening thresholds and precision only from one stage to the next, prune it to the feasible space \(\mathcal{S}\subseteq\mathcal{S}_{\mathrm{raw}}\).

\textbf{Multiplicative Space, Additive Analysis.}
The space grows as a product over components and stages and is far too large to search by execution. Yet all strategies draw their parts from the same \(\sum_{i}|\mathcal{C}_i|\) algorithm choices. Any property that can be analyzed per choice and then composed therefore costs \(\sum_{i}|\mathcal{C}_i|\) analyses rather than \(|\mathcal{S}|\) runs. This structure answers the first challenge, and Target-Scale Foresight is built on it.

\subsection{Target-Scale Foresight}
\label{sec:search}

Foresight rests on one separation: quantities that depend on system size are counted on the target or extrapolated toward it, and quantities that do not are measured once on small systems or kernels. Both the error and the cost models are built per algorithm choice and composed per strategy, so their cost follows the additive structure of \(\mathcal{S}\).

\textbf{Task-Aware Error Modeling.}
Approximation errors depend on molecular composition, local structure, and electron distribution, so they do not transfer across molecules. They do, however, arise locally: each approximation acts on products of nearby basis functions, atom-centered grid points, or screened shell quartets, and in gapped systems the density matrix decays exponentially with distance~\cite{kohn1996density,prodan2005nearsightedness}. Along a series of systems that share the local structure of the target, errors thus grow nearly linearly with size. \method therefore builds a series of small proxy systems matched to the target. On each proxy, it computes a tightly converged reference without further approximation, fixes the reference density, and substitutes the algorithm choice of one component at a time to measure its error, e.g., that of the auxiliary basis of RI-J, the grid of SGX, or the screening threshold of LinK.
Because approximations also change the self-consistent density, \method runs full SCF calculations for selected combinations and splits the signed error behind \(\Delta(\theta;x)\) into
\begin{equation}
\begin{aligned}
E_\theta[D_\theta]-E_{\mathrm{ref}}[D_{\mathrm{ref}}]
={}&
\underbrace{
E_\theta[D_\theta]-E_{\mathrm{ref}}[D_\theta]
}_{\text{fixed-density approximation error}}
\\
&+
\underbrace{
E_{\mathrm{ref}}[D_\theta]-E_{\mathrm{ref}}[D_{\mathrm{ref}}]
}_{\text{self-consistent density response}}.
\end{aligned}
\end{equation}
The density response, calibrated on these combinations, scales the sum of component errors, and the error of a schedule is that of its final stage. \method then fits the size dependence of each component error over the proxy series, validates the fit on held-out larger proxies, and attaches a prediction margin to obtain \(\widehat{\Delta}(\theta;x)\). We call this the \term{local-error model}.

\textbf{Hardware-Aware Performance Modeling.}
Gaussian integrals, matrix operations, and eigensolvers stress the hardware in different ways, and screening shifts their scaling with size, so neither aggregate complexity nor a \term{proxy-fitted model}, a power law fitted to timings on small proxies and extended to the target~\cite{calotoiu2013extrap}, reliably predicts target-scale runtime. \method instead separates the two sources of runtime. Size enters only through the workload \(W_k\) of each operation \(k\), which \method counts on the target system itself at a small fraction of the cost of running it; because integral screening depends on the density, it counts with the initial guess \(D_0\), and for LinK it enumerates the shell quartets that survive density-weighted screening~\cite{ochsenfeld1998link,kussmann2013preselective}. Hardware enters only through the throughput \(R_k\) of the general-purpose kernel, which does not depend on size and is measured by microbenchmarks on the target hardware per angular-momentum class~\cite{volkov2008benchmarking,jia2018dissecting}. The \term{W/R model} combines the two as \(\widehat{T}_k=W_k/R_k\), with one constant fitted on proxy systems for overheads a microbenchmark misses; eigensolvers~\cite{anderson1999lapack,marek2014elpa} are benchmarked directly on target-sized matrices with representative spectra. Summing over a strategy's stages, with iteration counts from the proxy systems and with data transfers and overlap, yields the predicted runtime \(\widehat{T}\) of its uncompiled program and that of each component, \(\widehat{T}_i\). The same counts give the size of every stored tensor and thus the predicted peak memory \(\widehat{M}\).

\textbf{Pareto-Guided Strategy Selection.}
Composing these per-choice predictions gives the error, runtime, and peak memory of every strategy in \(\mathcal{S}\) without evaluating any strategy separately. \method traces the predicted accuracy--runtime Pareto frontier~\cite{miettinen1998nonlinear} at the target scale, which shows the fastest strategy at every error level, and selects the fastest strategy \(\theta^\ast\) with \(\widehat{\Delta}\leq\varepsilon\) and \(\widehat{M}\leq M_{\max}\). It passes on \(\theta^\ast\), which fixes the algorithms, pairings, and schedule, together with the per-component runtimes \(\widehat{T}_i\) that guide compilation.

\subsection{Agent-Driven Compilation}
\label{sec:compile}

Compilation turns \(\theta^\ast\) into a program that runs faster than its uncompiled form. It changes only how \(\theta^\ast\) is computed, never what is computed: the implementation choices deferred in Sec.~\ref{sec:compose} are made here, while algorithms, parameters, and precision stay fixed, so \(\widehat{\Delta}\) still holds.

\textbf{Static Computation Graph.}
Once the molecule and \(\theta^\ast\) are fixed, the operators of every SCF stage, their dependencies, tensor shapes, and workloads are known before execution; only the sparsity of density-based screening varies, and it settles within the first iterations. Like a trained network frozen for deployment~\cite{chen2018tvm,ansel2024pytorch2}, the calculation forms a nearly static computation graph, unlike the general execution paths of a package, which must serve any molecule. This enables three kinds of ahead-of-time (AOT) optimization that a package cannot apply. Invariants are hoisted: quantities that depend only on the molecule and basis, such as the factorization of the overlap matrix~\cite{lowdin1950non}, are computed once rather than in every iteration. Kernels and solvers are specialized to the structure of the task, such as the band structure of the overlap matrix or the shell classes and tensor shapes of the basis~\cite{headgordon1988method,ufimtsev2008quantum}, with memory preallocated. And knowledge of the molecule is used directly, for example to tile the initial density from converged fragments~\cite{vanlenthe2006starting,gordon2012fragmentation}.

\textbf{Prediction-Guided Agentic Compilation.}
These optimizations are open-ended code transformations, such as a new solver form or a rewritten integral kernel, rather than fixed compiler passes, so coding agents carry them out~\cite{novikov2025alphaevolve,chen2025cupilot}. Because the runtime shares of components change with size, agents do not rank bottlenecks by small-scale timings; they receive the predicted target-scale runtime \(\widehat{T}_i\) of each component and work first on the components that dominate it. The microbenchmarks behind \(\widehat{T}_i\) also suggest how: a component limited by integral throughput calls for kernel specialization, one limited by dense linear algebra for a better solver form or data layout. For LinK at a fixed basis and threshold, for example, agents specialize the integral kernels to the angular-momentum classes of the basis and precompute the screened shell-pair lists.

\textbf{Validate, Measure, and Iterate.}
Each modification is first validated: component matrices and SCF energies on proxy systems must agree with those of the uncompiled program within a tolerance set well below \(\varepsilon\). Validated changes are then timed on the components they affect, and the measured component times \(T_i\) are updated; failing or slower changes are reverted. The loop ends when the budget \(B\) is spent or no remaining change pays off. The final program then runs once at the target scale, where its convergence, runtime, and peak memory are measured; its error remains the extrapolated \(\widehat{\Delta}\), which compilation leaves unchanged.

\section{Experimental Evaluation}
\label{sec:experiments}

\subsection{Experimental Setup}

\paragraph{Hardware Platforms.}
We evaluate \method on two single-node platforms: \textbf{(1)~CPU}, a Dell PowerEdge R7525 with two AMD EPYC 7513 processors (64 cores in eight NUMA domains) and 1\,TiB of memory; and \textbf{(2)~CPU--GPU}, a server with two AMD EPYC processors (64 cores), 512\,GB of memory, and one NVIDIA A100 SXM4 GPU with 80\,GB of device memory.

\paragraph{Baselines.}
For strategy coverage, we examine the algorithms, parameter options, and their composability in ORCA~6.1~\cite{neese2025orca6}, Gaussian~16, Q-Chem~6.3~\cite{epifanovsky2021qchem}, PySCF~2.15, and GPU4PySCF~1.9.1, based on public documentation. Because some commercial licenses restrict public performance benchmarking, we use the open-source PySCF~\cite{sun2020pyscf,sun2026pyscf} and GPU4PySCF~\cite{li2025gpu4pyscf,wu2025gpu4pyscf,li2026multigrid} as the CPU and GPU baselines for accuracy and performance, configured for the same physical model, orbital basis, error tolerance, and resources. The tuned baseline uses the fastest algorithms and settings the package supports that meet the error tolerance.

\paragraph{Metrics.}
We evaluate four aspects. \textbf{(1)~Strategy coverage}: the algorithms each package supports per component and how it combines them across components and SCF stages. \textbf{(2)~Numerical accuracy}: the total-energy error $\Delta E=|E-E_{\mathrm{ref}}|$ relative to a tightly converged RHF reference with the same geometry and basis. We use $\varepsilon=10^{-4}\,E_h$ (0.1\,m$E_h$), the accuracy scale of widely used approximations~\cite{weigend2006jkfit,helmich2021cosx}, as a representative error tolerance; each task may set its own. \textbf{(3)~Execution performance}: end-to-end SCF wall time and peak memory, including GPU memory where applicable. \textbf{(4)~Development efficiency}: total time to solution, including strategy analysis, code generation, debugging, validation, and the production run. All predictions are fixed before final evaluation, and timings are repeated under the same resource constraints.

\subsection{Dissecting \method}
\label{sec:dissect}

\runinbox{Task}{RHF/cc-pVTZ~\cite{dunning1989gaussian} on the linear alkane C$_{862}$H$_{1726}$ (2{,}588 atoms, 50{,}024 basis functions) on the CPU platform with $\varepsilon=10^{-4}\,E_h$. Shorter alkanes, from C$_{2}$H$_{6}$ to C$_{172}$H$_{346}$, serve as proxy systems. All experiments in this section use this task.}

\begin{figure}[t]
\centering
\includegraphics[width=\linewidth]{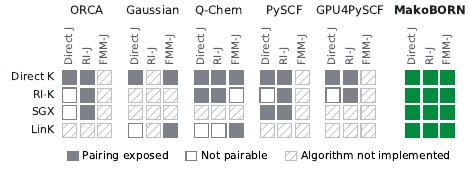}
\caption{$(J,K)$ algorithm pairings each package exposes for RHF, per public documentation (ORCA~6.1, Gaussian~16, Q-Chem~6.3, PySCF~2.15, GPU4PySCF~1.9.1).}
\label{fig:space}
\end{figure}

\qpart[sec:space]{Strategy Coverage Beyond Existing Packages}{Does \method break the strategy limits of general-purpose packages?}{Yes. \method unfolds the algorithms, pairings, and schedules that each package fixes into a strategy space far beyond the reach of any single package.}
We focus on J and K construction, which dominate both the cost and the approximation error of RHF.
According to their documentation, existing packages are restricted at each of the three levels of Sec.~\ref{sec:compose}.
\lvl{Algorithms}Each algorithm comes in a largely fixed form: variants are rare (only Q-Chem offers local and pair-atomic RI), parameters come in a few preset levels (e.g., the DEFGRID1--3 grids of ORCA~\cite{helmich2021cosx}), and most packages run in double precision only. \method searches variants and parameters freely.
\lvl{Pairings}Algorithms are exposed only as a few preset pairings (Fig.~\ref{fig:space}), such as RIJCOSX in ORCA, clustered around each package's design focus: RI-J-based pairings in ORCA, PySCF, and GPU4PySCF, and FMM-J with LinK in Gaussian and Q-Chem. Choosing a package thus largely fixes the strategy family; \method pairs any J algorithm with any K algorithm.
\lvl{Schedules}One pairing and one set of grids, thresholds, and precision apply to the entire SCF run, although early iterations tolerate looser settings. \method switches pairings and settings between SCF stages.

\begin{figure}[t]
\centering
\includegraphics[width=\linewidth]{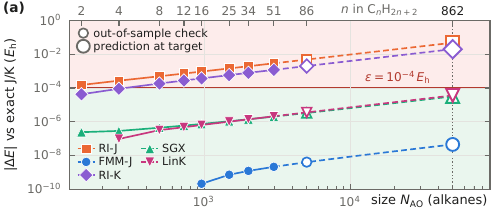}\\[2pt]
\includegraphics[width=\linewidth]{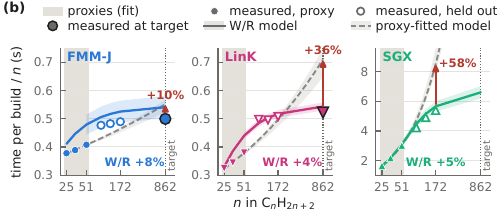}
\caption{Target-scale foresight on alkanes (cc-pVTZ). \textbf{(a)}~Energy error of each J/K approximation. \textbf{(b)}~Build time per carbon; red arrows: error of the proxy-fitted model.}
\label{fig:pred}
\end{figure}

\qpart[sec:predict]{Target-Scale Foresight for Strategy Selection}{Does \method foresee target-scale error and runtime accurately enough to select the most promising strategy without target-scale runs?}{Yes. \textit{Error:} models fitted on proxies predict held-out errors to within 2\%. \textit{Runtime:} the W/R model predicts build times to within 8\% up to the target, where the proxy-fitted model misses by up to 79\%. \textit{Selection:} the resulting Pareto frontier selects a strategy predicted at 5.6\,h, nearly $20\times$ faster than the best baseline strategy below $\varepsilon$.}

\lvl{Accuracy}Error can be fitted on proxies and extended to the target, because it accumulates locally, with a near-constant increment per carbon atom. Local-error models, linear in $n$ and fitted up to C$_{51}$H$_{104}$, predict the held-out C$_{86}$H$_{174}$ to within 2\% for all five approximations (Fig.~\ref{fig:pred}a). At the target, RI-J and RI-K with standard auxiliary bases exceed $\varepsilon$ by more than two orders of magnitude, whereas FMM-J, SGX, and LinK stay below it.

\lvl{Runtime}Runtime cannot be extended this way: screening shifts its scaling exponent with size.
\method therefore counts each kernel's workload on the molecule itself, the target included (W/R model, Sec.~\ref{sec:search}).
The resulting build times stay within 8\% of the measured ones at every held-out size, and at the farthest measurement of each kernel they are off by 4\% (LinK) and 8\% (FMM-J) at C$_{862}$H$_{1726}$ and by 5\% (SGX) at C$_{172}$H$_{346}$ (Fig.~\ref{fig:pred}b).
At the same points, the proxy-fitted model, fitted to the same timings, is off by 36\%, 10\%, and 58\%, and by up to 44\%, 14\%, and 79\% with other fit windows.
In addition, diagonalization is benchmarked directly at the target dimension rather than counted, and lies within 4\% of the production run.

\begin{figure}[t]
\centering
\includegraphics[width=\linewidth]{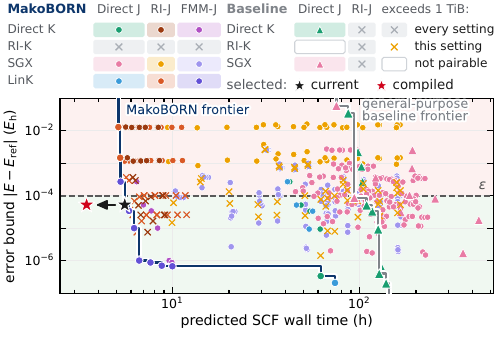}
\caption{Predicted accuracy--runtime Pareto frontiers for RHF/cc-pVTZ on C$_{862}$H$_{1726}$ (CPU platform). Each point is one strategy, colored by its J--K pairing (legend grids); circles: \method, triangles: baseline.}
\label{fig:landscape}
\end{figure}

\lvl{Selection}Together, the two predictions place every candidate on an accuracy--runtime plane at the target scale before any target-scale run (Fig.~\ref{fig:landscape}).
The Pareto frontier of this plane gives the fastest strategy at each error level, and its fastest point below $\varepsilon$ that fits in the 1\,TiB node memory is the selection: FMM-J with LinK (screening threshold $3\times10^{-8}$), predicted at 5.6\,h.
The general-purpose baseline frontier, built from the strategies PySCF exposes, reaches $\varepsilon$ only at 109\,h, nearly $20\times$ later.

\begin{figure}[t]
\centering
\includegraphics[width=\linewidth]{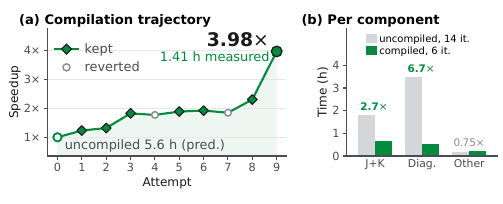}
\caption{Agent-driven compilation of the selected strategy (C$_{862}$H$_{1726}$). \textbf{(a)}~Speedup of each attempt over the uncompiled program. \textbf{(b)}~Time of the whole calculation by component.}
\label{fig:synthesis}
\end{figure}

\qpart[sec:synth]{Agent-Driven Compilation of the Selected Strategy}{Does \method compile the selected strategy to run much faster without loss of accuracy?}{Yes. Exploiting the static computation graph, \method makes it run 3.98$\times$ faster at an unchanged error.}
Starting from the uncompiled program, which implements the selected strategy with general-purpose kernels and solvers, agents keep 7 attempts and cut the whole calculation from the predicted 5.6\,h to 1.41\,h, measured end to end, at an unchanged error (3.98$\times$; Fig.~\ref{fig:synthesis}a, Table~\ref{tab:e2e}).
The largest gains come from what a general package cannot fix in advance. Since $S$ is fixed, it is factorized once and its band structure replaces the dense transform (attempts~1, 2, 8), making diagonalization, 64\% of the uncompiled time, 6.7$\times$ faster (Fig.~\ref{fig:synthesis}b). Since the molecule is known, the initial density is tiled from its converged fragments (9), cutting the SCF from 14 to 6 iterations.

\begin{table}[t]
\centering
\footnotesize
\setlength{\tabcolsep}{1.5pt}
\renewcommand{\arraystretch}{1.1}
\begin{tabular*}{\linewidth}{@{\extracolsep{\fill}}l c c c c c c@{}}
\toprule
 & $\Delta E$ & Total & Mem. & Dev. & \multicolumn{2}{c}{Speedup} \\
\cmidrule(l){6-7}
Configuration & (m$E_h$) & (h) & (GB) & (h) & w/o Dev. & w/ Dev. \\
\midrule
Baseline, tuned$^\dagger$ & 0.052 & 109 & 700 & 0    & 1.0$\times$ & 1.0$\times$ \\
\method, uncompiled$^\dagger$ & 0.052 & 5.6 & 418 & 3.8  & 19$\times$ & 12$\times$ \\
\rowcolor{yesgreen!10}
\method, compiled     & 0.053 & 1.41 & 334 & 9.7  & 77$\times$ & 9.8$\times$ \\
\bottomrule
\end{tabular*}
\caption{End-to-end results for C$_{862}$H$_{1726}$. Dev.: cumulative development time. $^\dagger$Predicted values (Sec.~\ref{sec:predict}); Baseline cannot run at this size, as its direct J/K fails once $n_\mathrm{AO}^2>2^{31}$.}
\label{tab:e2e}
\end{table}

\qpart[sec:e2e]{End-to-End Gain and Development Cost}{Is the end-to-end gain worth the development cost?}{Yes. \method runs about 77$\times$ faster than the tuned baseline below $\varepsilon$, and still roughly 9.8$\times$ faster for a single run with its development included.}
Strategy selection alone makes the calculation about 19$\times$ faster than the tuned baseline, and compilation raises this to about 77$\times$, all below $\varepsilon$ (Table~\ref{tab:e2e}).
Development, from strategy selection to the validated program, takes 9.7\,h; added to one run, it still leaves a speedup of roughly 9.8$\times$.
The gain thus pays off from the first run and grows with each further one.

\subsection{Large-Scale Biomolecular Calculations}
\label{sec:cases}

With existing packages, neither calculation is practical on a single node with one GPU. For ubiquitin, general-purpose quantum chemistry packages take days, and Mako~\cite{han2025mako} brings the time to under 1\,h only by spreading the calculation over 64 GPUs. For the B-DNA duplex, with nearly 50{,}000 atomic orbitals, the general-purpose route is predicted to take about 1{,}400\,h.

\runinbox{Tasks}{Two biomolecular RHF calculations.\par
\vspace{4pt}%
{\footnotesize\setlength{\tabcolsep}{3pt}\renewcommand{\arraystretch}{1.08}%
\begin{tabular*}{\linewidth}{@{\extracolsep{\fill}}c c@{}}
\multicolumn{2}{@{}l@{}}{\makebox[\linewidth][l]{%
  \sbox0{1{,}231 atoms, 22{,}442 basis fns.}%
  \makebox[\wd0]{\includegraphics[height=0.40in]{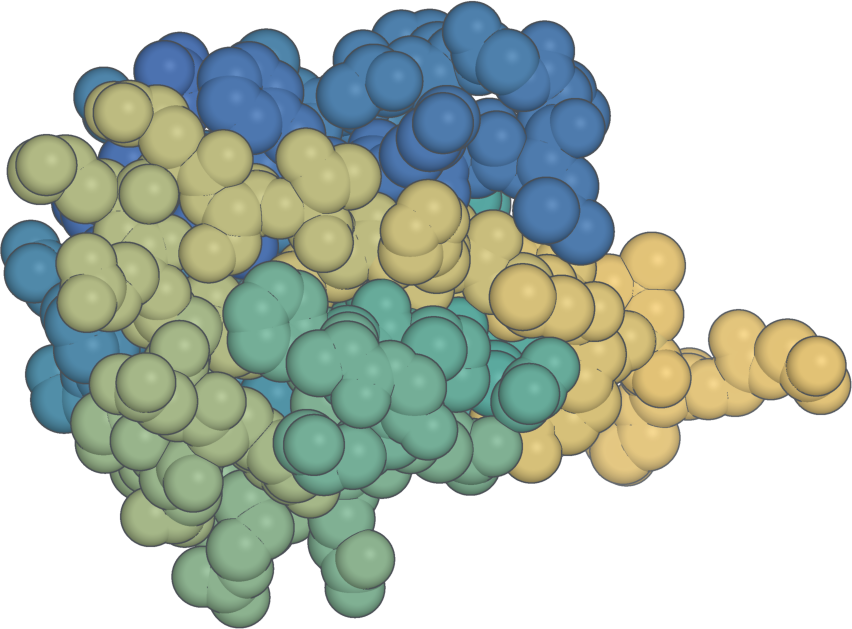}}\hfill
  \includegraphics[width=1.7in]{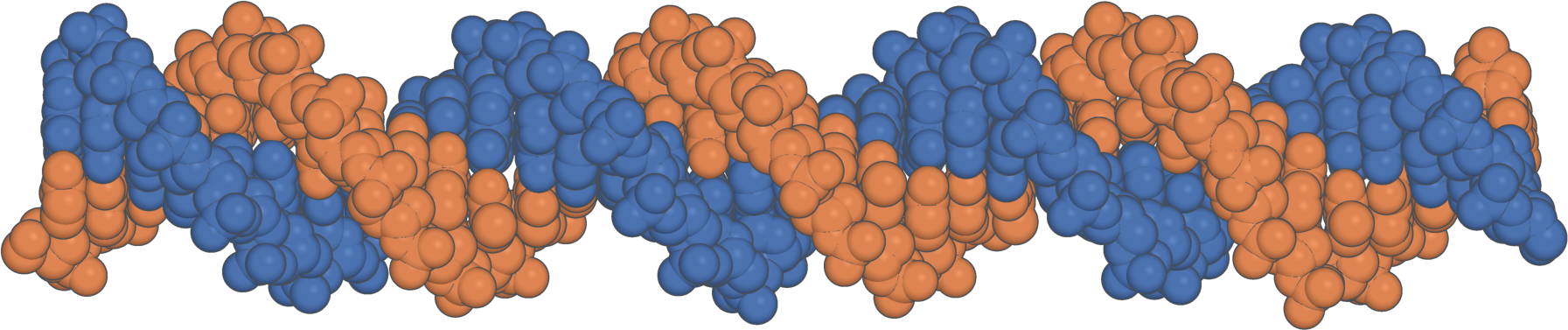}}} \\
\addlinespace[3pt]
\textbf{Ubiquitin} (PDB 1UBQ~\cite{vijaykumar1987ubiquitin}) & \textbf{B-DNA duplex}~\cite{ni2021cim} \\
\textcolor{black!60}{$4.3\times3.0\times2.7$\,nm} & \textcolor{black!60}{$11.6\times2.3\times2.3$\,nm} \\
1{,}231 atoms, 22{,}442 basis fns. & 2{,}282 atoms, 49{,}825 basis fns. \\
\addlinespace[2pt]
\arrayrulecolor{black!25}\midrule\arrayrulecolor{black}
\multicolumn{2}{@{}c@{}}{\textit{Both tasks:}\ def2-TZVP basis~\cite{weigend2005balanced}\ $\cdot$\ CPU--GPU\ $\cdot$\ $\varepsilon=10^{-4}\,E_h$} \\
\end{tabular*}}}

\qbox{Can \method bring large-scale biomolecular calculations within practical reach?}{Yes. \method runs ubiquitin on one GPU instead of Mako's 64 GPUs, and the B-DNA duplex in 8.1\,h instead of the baseline's predicted 1{,}400\,h ($173\times$), both below $\varepsilon$.}

\lvl{Strategy}The best strategy changes with the task (Table~\ref{tab:strategies}). The alkane runs on the CPU alone, where its chain lets LinK screening discard most integrals; ubiquitin also has a GPU, so exchange switches to SGX, which runs on it (LinK would take a predicted 10\,h). From ubiquitin to the B-DNA duplex, the auxiliary functions grow from about 233{,}000 to about 628{,}000, and memory decides the Coulomb choice. For ubiquitin, FMM-J fits only the near field, so the three-center integrals fit in host memory, whereas full-space density fitting would not. For the B-DNA duplex, the dense Coulomb metric reaches 3.2\,TB, far beyond the memory of a single node; \method instead fits in the attenuated metric $\operatorname{erfc}(\omega r)/r$, whose Cholesky factor is banded along the helix and takes 142\,GB. A robust energy expression keeps the resulting fitting error second order. Shape also changes the accuracy demand: the SGX error accumulates along the helix, so the duplex needs a 434-point grid where ubiquitin needs 302. Together, these choices lie beyond existing packages: none offers the FMM-J--SGX pairing of ubiquitin or a two-stage SCF schedule.

\begin{table}[t]
\centering
\footnotesize
\setlength{\tabcolsep}{2pt}
\renewcommand{\arraystretch}{1.08}
\providecommand{\sidelabel}[3]{\multirow{-#1}{*}{\rotatebox[origin=c]{90}{\textcolor{#3}{\textbf{\textit{#2}}}}}}%
\providecommand{\bandcell}[2]{\multicolumn{1}{@{}>{\columncolor{#1}[0pt][1pt]}c@{\hspace{4pt}}}{#2}}%
\providecommand{\taskc}[1]{\bandcell{black!10}{#1}}\providecommand{\stratc}[1]{\bandcell{yesgreen!18}{#1}}%
\begin{tabular*}{\linewidth}{@{}c@{\hspace{4pt}}l@{\extracolsep{\fill}}c c c@{}}
\toprule
 & & Alkane & Ubiquitin & B-DNA duplex \\
\midrule
\taskc{} & Shape                 & Chain    & Globular  & Double helix \\
\taskc{} & Basis functions       & 50{,}024 & 22{,}442  & 49{,}825 \\
\taskc{\sidelabel{3}{Task}{black!70}}
       & Platform              & CPU      & CPU--GPU  & CPU--GPU \\
\midrule
\stratc{} & Coulomb (J)          & FMM-J    & FMM-J     & RI-J \\
\stratc{} & \quad near field      & Exact    & RI (233k) & RI (628k) \\
\stratc{} & \quad far field       & Multipole & Multipole & --- \\
\stratc{} & \quad setting         & $R{=}20$, $p{=}10$, $\tau$ & $\theta{=}0.7$, $p{=}12$ & robust erfc, $\omega{=}2$ \\
\stratc{} & Exchange (K)         & LinK     & SGX       & SGX \\
\stratc{} & \quad stage 1        & ---      & $n_\Omega{=}110{\to}194$ & $n_\Omega{=}110{\to}194$ \\
\stratc{\sidelabel{7}{Strategy}{yesgreen!65!black}}
        & \quad stage 2        & $\tau{=}3{\times}10^{-8}$ & $n_\Omega{=}302$ & $n_\Omega{=}434$ \\
\bottomrule
\end{tabular*}
\caption{{Strategy that \method selects for each task.} SCF stage~1 converges to $10^{-5}E_h$ with a smaller auxiliary basis, stage~2 (listed) to $10^{-8}E_h$; parentheses: auxiliary functions. J: near/far boundary $R$ ($a_0$) or $\theta$; multipole order $p$; $\omega$ ($a_0^{-1}$): attenuation of the robustly fitted erfc metric (other RI: Coulomb metric); the LinK threshold $\tau$ also screens the near field. K: $n_\Omega$, Lebedev points per shell.}
\label{tab:strategies}
\end{table}

\begin{figure}[t]
\centering
\includegraphics[width=\linewidth]{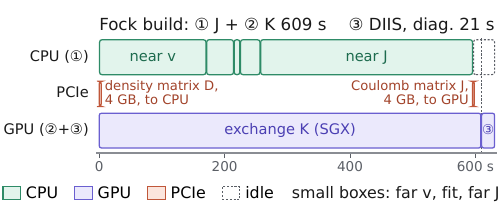}
\caption{One SCF iteration of the compiled ubiquitin program.}
\label{fig:ubq-allcpu}
\end{figure}

\lvl{Program}We take ubiquitin as the example (Fig.~\ref{fig:ubq-allcpu}). Under the selected strategy, the SGX exchange matrix K dominates each iteration and runs on the GPU, whereas the three-center integrals of FMM-J fit in host memory, so the Coulomb matrix J can move entirely to the CPU. The program that \method compiles computes the two concurrently: the devices exchange data only twice, at the start and end of the Fock build, J finishes within the time of K, and the otherwise idle CPU is put to work. Mainstream packages provide no such overlap of J and K across devices, because whether it is feasible, and how to split the work, depend on tensor sizes and workloads known only once the molecule and strategy are fixed. \method compiles for exactly such a fixed task, so the split becomes just one more implementation choice for its agents to explore (Sec.~\ref{sec:compile}).

\begin{figure}[t]
\centering
\includegraphics[width=\linewidth]{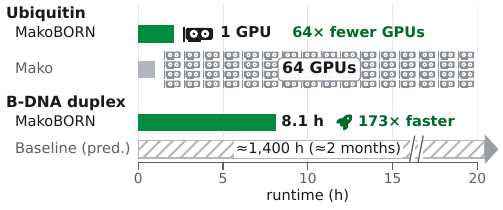}
\caption{Runtime of the compiled program on one A100 against Mako on 64 GPUs (ubiquitin) and the predicted baseline (B-DNA duplex).}
\label{fig:cases}
\end{figure}

\lvl{Result}Both calculations complete on a single node with one A100 (Fig.~\ref{fig:cases}). Ubiquitin takes 2.1\,h; Mako~\cite{han2025mako} needs under 1\,h but uses 64 GPUs, so \method reaches the same order of runtime with 64$\times$ fewer GPUs. The B-DNA duplex takes 8.1\,h, against about 1{,}400\,h (about two months) predicted for the baseline, a 173$\times$ speedup. The measured errors, 5.3 and 3.0\,$\times10^{-5}E_h$, stay below $\varepsilon$. Development, from strategy selection to the validated program, takes 8.5 and 12.6\,h, far less than the time saved on the B-DNA duplex, so building the program pays off within a single calculation.

\section{Future Work and Conclusion}
\label{sec:conclusion}

\paragraph{Future Work.}
Beyond single-node Hartree--Fock, three directions remain open. First, \method can be extended to more physical models in quantum chemistry. Second, foresight rests on controllable accuracy and predictable cost, which call for further study of error control and prediction, for example uncertainty propagation through SCF calculations~\cite{diazfarias2026qcup}. Third, the programs \method builds are irregular yet static, with their structure known before execution; mapping such scientific algorithms onto complex hardware topologies calls for compilers dedicated to scientific computing.

\paragraph{Conclusion.}
We presented \method, a foresighted agentic framework that synthesizes task-specific quantum chemistry programs on demand. By predicting each strategy's target-scale error and runtime before implementation, \method makes program development a budgeted part of the calculation and brings calculations beyond general-purpose software within practical reach. We hope it moves scientific computing from adapting problems to existing programs toward building a program for each task.

\clearpage
{\small\bibliographystyle{ieeenat_fullname_doi}\bibliography{main}}
\end{document}